\documentclass[letterpaper]{article} 
\usepackage{aaai23}  
\usepackage{times}  
\usepackage{helvet}  
\usepackage{courier}  
\usepackage[hyphens]{url}  
\usepackage{graphicx} 
\usepackage{natbib}  
\usepackage{caption} 
\usepackage{algorithm}
\usepackage{algorithmic}

\usepackage{newfloat}
\usepackage{listings}
\DeclareCaptionStyle{ruled}{labelfont=normalfont,labelsep=colon,strut=off} 
\floatstyle{ruled}
\newfloat{listing}{tb}{lst}{}
\floatname{listing}{Listing}
\usepackage{booktabs}       
\usepackage{nicefrac}       
\usepackage{url}            
\usepackage{amsfonts}
\usepackage{tabularx}
\usepackage{graphics}
\usepackage{mdwlist}
\usepackage{xspace}
\usepackage{amsmath}
\usepackage{mathtools}
\usepackage{wrapfig}
\usepackage{makecell}

\newcommand{\fullName}{Epidemiologically-informed Neural Networks\xspace}
\newcommand{\ourmethod}{\textsc{EINNs}\xspace}

\newcommand{\covid}{COVID-19\xspace}

\newcommand{\rnn}{\textsc{RNN}\xspace}
\newcommand{\ode}{\textsc{SEIRM}\xspace}
\newcommand{\odeflu}{\textsc{SIRS}\xspace}
\newcommand{\generation}{\textsc{Generation}\xspace}
\newcommand{\regularization}{\textsc{Regularization}\xspace}
\newcommand{\ensembling}{\textsc{Ensembling}\xspace}
\newcommand{\pinn}{\textsc{PINN}\xspace}
\newcommand{\problem}{\textsc{Incorporating Epi-Dynamics in NNs}\xspace}

\title{EINNs: Epidemiologically-Informed Neural Networks}
\author{
Alexander Rodr\'iguez$^1$,
Jiaming Cui$^1$,
Naren Ramakrishnan$^{2}$,
Bijaya Adhikari$^3$, {\normalfont and}
\\B. Aditya Prakash$^1$
}
\affiliations{
    $^1$College of Computing, Georgia Institute of Technology, Atlanta, USA\\
    $^2$Department of Computer Science, Virginia Tech, Arlington, USA\\
    $^3$Department of Computer Science, University of Iowa, Iowa City, USA\\
    \{arodriguezc, jiamingcui1997, badityap\}@gatech.edu, naren@cs.vt.edu, bijaya-adhikari@uiowa.edu
}

\usepackage{bibentry}

\begin{document}

\maketitle

\begin{abstract}
We introduce EINNs, a framework crafted for epidemic forecasting that builds upon the theoretical grounds provided by mechanistic models as well as the data-driven expressibility afforded by AI models, and their capabilities to ingest heterogeneous information. Although neural forecasting models have been successful in multiple tasks, predictions well-correlated with epidemic trends and long-term predictions remain open challenges. Epidemiological ODE models contain mechanisms that can guide us in these two tasks; however, they have limited capability of ingesting data sources and modeling composite signals. Thus, we propose to leverage work in physics-informed neural networks to learn latent epidemic dynamics and transfer relevant knowledge to another neural network which ingests multiple data sources and has more appropriate inductive bias.
In contrast with previous work, we do not assume the observability of complete dynamics and do not need to numerically solve the ODE equations during training. Our thorough experiments on all US states and HHS regions for COVID-19 and influenza forecasting showcase the clear benefits of our approach in both short-term and long-term forecasting as well as in learning the mechanistic dynamics over other non-trivial alternatives.
\end{abstract}

\section{Introduction}
\label{sec:intro}
The COVID-19 pandemic has led to a maturing of methods for epidemic modeling and forecasting with the CDC establishing the first Center for Forecasting and Outbreak Analytics in 2021.
A variety of forecasting
innovations in machine learning and deep learning were developed--e.g.,~\cite{rodriguez_deepcovid_2021,kamarthi_when_2021}--with many lessons learned for COVID-19 and future pandemics. As the current experience has shown, predicting and preventing epidemics is one of the major challenges with far reaching impacts on health, economy and broad social well being\footnote{www.who.int/activities/preventing-epidemics-and-pandemics}\!.

From this perspective, active participation by several academic and industrial teams (including by coauthors) in multiple CDC-led forecasting initiatives has led to two broad themes that are important for epidemic modeling. First, modern disease surveillance has grown by leaps and bounds yielding novel data sources that can shed light into happenings real-time. \emph{Statistical/ML epidemic models} leverage these data sources to provide dramatic improvements in short-term forecasting (usually 1-4 weeks ahead). At the same time, as these methods do not explicitly learn mechanistic dynamics, such methods do not provide understanding of how the epidemic will unfold at even longer time horizons, and do not support posing causal and counterfactual questions (e.g., design of countermeasures). Such longer-term forecasting remains the province of \emph{mechanistic epidemic models} that can support scenario-based understanding of epidemic progression (e.g., "what will happen if schools are closed?").  
However, these methods present scalability issues, their calibration is prone to noise~\cite{hazelbag2020calibration}, and they have limited capability to ingest multimodal data sources~\cite{viboud2019future}. 
At the intersection of these two modeling approaches, we have hybrid models that make compartmental models (based on ordinary differential equations -- ODEs) more informed of these data sources~\cite{shaman_real-time_2013,arik_interpretable_2020}. 
However, most of these approaches use the mechanistic model for prediction, thus, they are not flexible enough to fit the complex patterns in epidemics and have very few tunable parameters. In addition, they are not easily generalizable to new models/data sources or do not aim to incorporate ODE dynamics from first principles (e.g. predict ODE parameters instead of the ODE states).

\begin{figure}[t!]
    
    \centering
    \small
    \begin{tabular}{cc}
    \centering
    \includegraphics[width=0.46\linewidth]{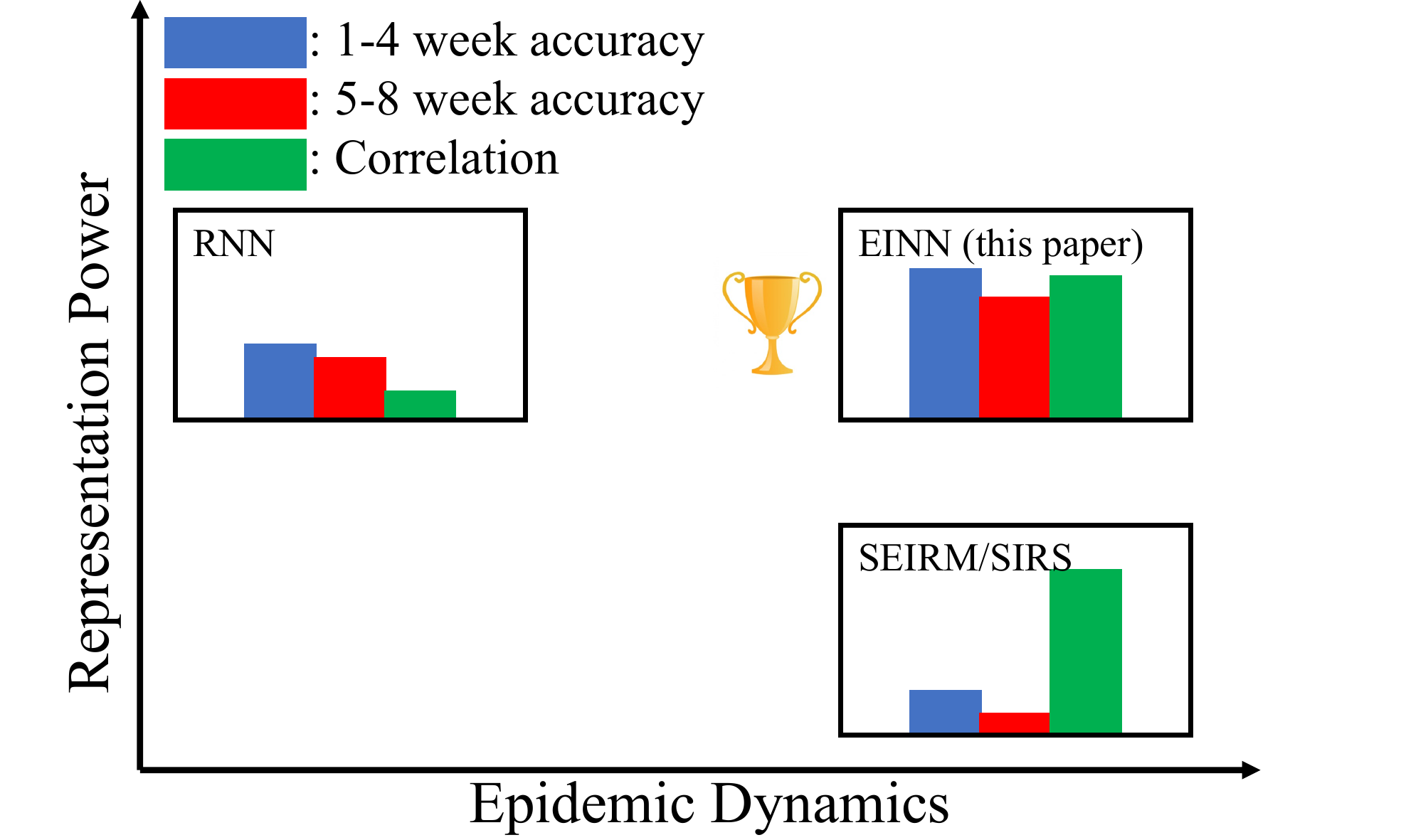} & \hspace{-0.3cm}
    \includegraphics[width=0.48\linewidth]{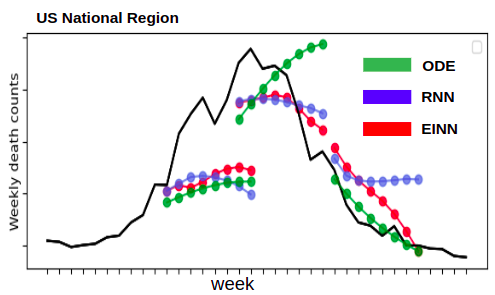} \\
    
    (a) Modeling spectrum & (b) Examples of   \\
    and performance &  forecasts \\
  \end{tabular}
    \caption{
   Our method, \ourmethod, (a) takes the best from both modeling approaches and is suitable for short- and long-term forecasting, and (b) its predictions are well-correlated with the epidemic trends. 
    \vspace{-0.5cm}
    }
    \label{fig:teaser}
\end{figure}

In this paper, we develop a \textit{general framework} for incorporating epidemic dynamics from a mechanistic model into a neural framework for forecasting, which enables seamless integration of multimodal data, greater representation power, and inclusion of composable neural modules of learned representations. 
Our focus is to leverage the selective superiorities of both approaches (see Figure~\ref{fig:teaser}) to have predictions that are \emph{accurate} (low error) and \emph{well-correlated} with even longer-term epidemic trends, than what is usually studied in past literature.

Recent lines of research (e.g.,~\cite{karniadakis_physics-informed_2021}) aim to bridge scientific models (usually represented as differential equations) and ML algorithms. Specifically, the rapidly growing literature in physics-informed neural networks (PINNs) has demonstrated that integrating neural networks with ODEs can lead to large scalability improvements as forward passes (over the ODE) are no longer needed~\cite{lu_deepxde_2020}. In addition, the neural network flexibility and gradient-based learning enables robust solutions in spite of noise~\cite{yang2021b}. 
We propose to build upon this body of work to incorporate the dynamics of a mechanistic epidemic model into deep neural models. Our goal requires innovations to the PINN literature as many of the compartments in epi-models are latent (e.g. the actual number of people exposed to the disease) while most work in PINNs has only experimented with all states of the system dynamics being observable.

In addition, PINNs are limited to working with the variables that are described in the mechanistic equations which limits their capabilities to ingest data sources.
On top of that, PINNs often use a simple multi-layer perceptron architecture whose inductive bias is often insufficient for sequential data.
The main technical innovation of this paper comes from designing a transfer learning framework for transferring learned dynamics from a PINNs to a Recurrent Neural Network (RNN) that can ingest exogenous data features (data not-represented in the ODE). The goal of this is to have an RNN aware of epidemic ODE dynamics that has learned how data features (inputs to the RNN) shape/connect to the latent epidemic dynamics (outputs of the RNN).

We summarize our contributions as follows: 
\par$\bullet$ \textbf{Push the boundaries of data-driven epi forecasting via integration of ODE-based mechanistic dynamics:} 
We introduce \fullName (EINNs), a new framework to bridge the gap between mechanistic and neural models for epidemiology. Our method incorporates epidemiological (expert) knowledge embedded in ODE models into neural networks (NNs) that ingest heterogeneous sources of data. 
The key idea in EINNs is utilizing a PINN to learn the latent epidemic dynamics and transfer its learned representations to another neural model with more appropriate inductive bias (RNN) and capable of ingesting heterogeneous sources of data that are exogenous to the ODE equations.
\par$\bullet$ \textbf{Transfer learning via gradient matching of dynamics:} We propose a novel method to transfer learned representations from a PINN (source model) to another neural network (target model). This is based on matching the gradients of mechanistic ODEs in the target model as we do so when training the source model. This cannot be directly done in the target model (RNN) due to its neural architecture and data inputs. Therefore, we propose approximating internal representations of source and target models to enable the target model to learn to match the ODE gradient.
\par$\bullet$ \textbf{Extensive empirical evaluation:} We evaluate our method in the challenging task of weekly COVID-19 forecasting (in 48 geographical regions and a period of 8 months) and flu (in 10 regions and 5 months). Our results showcase that our method can indeed leverage the `best of both worlds' compared to other non-trivial ways of merging such approaches. 
We believe this opens new venues for exploring how AI can better complement domain knowledge in traditional epidemiology. 

\section{Related Work}
\label{sec:related_work}
\par\noindent\textbf{\underline{Mechanistic and ML models for epidemic forecasting}}
Epidemic mechanistic models~\cite{hethcote_mathematics_2000} like the popular SIR are designed using domain knowledge of the epidemic dynamics. They model causal underpinnings to explain empirically observed variables (e.g., mortality), and ODE-based models have been a workhorse of epidemiology since the late 18th century~\cite{marathe_Comp._2013}.
More recently, there are several successful applications of ML to short-term forecasting~\cite{osthus_dynamic_2019,brooks_nonmechanistic_2018,adhikari_epideep:_2019,rodriguez_deepcovid_2021} which led them to be often ranked among the top performing models in these tasks~\cite{cramer_evaluation_2021,reich_collaborative_2019}. Some of the recent deep learning innovations include advances in incorporting multi-view and multimodal data~\cite{kamarthi_camul_2021}, spatial correlations~\cite{deng_cola-gnn_2020,jin_interseries_2021}, transfer learning for domain adaptation~\cite{rodriguez_steering_2021} and  non-parametric approaches~\cite{kamarthi_when_2021,zimmer_influenza_2020}.

\par\noindent\textbf{\underline{Hybrid epidemic models}}
They integrate mechanistic models and ML approaches~\cite{rodriguez2022data}. 
Some lines of work use statistical techniques to estimate the mechanistic parameters (e.g. transmission rate)~\cite{arik_interpretable_2020,wang2021bridging}, learn from simulation-generated data~\cite{wang_defsi_2019}, or use the ODEs as regularization~\cite{gao_stan_2021}. 
However, in addition to the previously mentioned flexibility problems, these models require a forward pass over the mechanistic models, which may become very expensive for long periods and large mechanistic models. Furthermore, they often need to discretize the continuous ODE space, which is a delicate process as it has been found that sub-optimal discretization can impede the learning process~\cite{thuerey2021pbdl}.
Our approach utilizes PINNs which allows skipping forward passes and discretization.
Perhaps the most prominent in this line of research is the work by~\cite{shaman_real-time_2013} that integrated search volume data into an SIRS ODE model via the principle of data assimilation. 
However developing such data assimilation approaches requires a large amount of domain knowledge and cannot be used to incorporate many of the data features studied in our work (e.g., mobility). 

\par\noindent\textbf{\underline{Physics-informed neural networks (PINNs)}}
PINNs are universal function approximators via neural networks that embed knowledge of ODEs and partial differential equations (PDEs) via unsupervised loss functions based on these equations.
They have been used for forward and inverse problems with ODEs in a variety of domains including computational biology~\cite{yazdani_systems_2020}.
PINNs have connections to implicit neural representations~\cite{sitzmann2020implicit} in the sense that they both provide continuous representations breaking the discretization (grid) limitation, which is advantageous for when data samples are irregular.
Previous work often use a multi-layer perceptron architecture because they are amenable for direct computation of derivatives of neural network outputs with respect to its inputs (via autograd). Indeed, incorporating inductive biases into PINNs is an active research area--e.g., convolutional layers \cite{wandel2022spline} and graph neural networks~\cite{kumar2021grade}--and to our best knowledge a recurrent neural architecture for PINNs remains an open problem.
Also, incorporating exogenous variables to this framework and working with partially observable systems are largely unexplored problems~\cite{cai2021deepm,wang_understanding_2021}.
Our approach \ourmethod extends the capabilities of PINNs by directly addressing these limitations in the context of epidemiology. 

A remotely related but popular line of work for learning dynamical systems is neural ODE~\cite{chen_neural_2018}.
While PINNs learn from the system dynamics represented in an ODE or PDE (which represent domain expert knowledge from epidemiologists), neural ODEs learn an unknown ODE via neural networks and do a continuous modeling using a numeric ODE solver. Since neural ODEs cannot incorporate domain ODEs, they are not applicable to our problem.

\section{Background} 
As mentioned earlier, we aim on merging neural models with epidemiological dynamics from a mechanistic model. Without loss of generality, here we introduce instantiations of such models which we refer to as building blocks. Their definitions help us to explain the formulation of the problem \problem{} and later our implementation and experiments. Additionally, we briefly introduces PINNs as formulated for Systems Biology.

\subsection{ODE-based mechanistic epidemic models}
\label{subsec:seirm}

Our first building block is a mechanistic epidemic model.
Epidemiologists use different mechanistic models for each disease because infection dynamics and disease progression varies~\cite{hethcote_mathematics_2000}. 
In this paper we use COVID-19 and influenza as a vehicle to demonstrate the benefits of our general framework, therefore, we use two different mechanistic epidemic models: SEIRM and SIRS (SIRS description is similar to SEIRM and can be found in our appendix). 

\noindent\par\textbf{\underline{SEIRM model for COVID-19}} 
The SEIRM ODE model consists of five compartments: Susceptible ($S$), Exposed ($E$), Infected ($I$), Recovered ($R$), and Mortality ($M$). 
It is parameterized by four variables $ \Omega = \{ \beta, \alpha,  \gamma, \mu\}$, where $\beta$ is the infectivity rate, $1/\alpha$ is the mean latent period for the disease,
$1/\gamma$ is the mean infectious period,
and $\mu$ is the mortality rate. Due to COVID-19's prolonged incubation period, the SEIRM has been broadly used in modeling its progression~\cite{wu2020nowcasting,morozova2021one}. It has also been used by the CDC in modeling transmission of Ebola~\cite{gaffey2018application}. 
To capture the evolving nature of the dynamics and spread of COVID-19 (e.g. consider the multiple variant waves), we leverage the dynamic version of the SEIRM model, where the parameters governing the disease progression themselves evolve over time. In such a setting, the dynamics is governed by the set of parameters $ \Omega_{t} = \{ \beta_t, \alpha_t,  \gamma_t, \mu_t\}$  at the given time-stamp $t$. Let $\textbf{s}_{t} = [ S_t, E_t, I_t, R_t, M_t ]^T$ be the values taken by the states at time $t$. (ODE state $S_t$ represents the number of susceptible people at time t, similar for the other states). Then, the ODEs describing the SEIRM model are given by $f_{\text{\tiny ODE}}(\textbf{s}_{t}, \Omega_t)$:
\begin{gather*}
\label{eq:seirm}
\frac{dS_t}{dt} = - \beta_t \frac{S_t I_t}{N} \qquad \frac{dE}{dt}  = \beta_t \frac{S_t I_t}{N} - \alpha_t E_t \\\nonumber 
\frac{dI_t}{dt} = \alpha_t E_t  - \gamma_t I_t - \mu_t I_t \qquad \frac{dR_t}{dt} = \gamma_t I_t \qquad \frac{dM_t}{dt} = \mu_t I_t.
\end{gather*}
In our SEIRM model, only mortality is considered observed~\cite{wu2020nowcasting}, therefore, ODE states $S_t, E_t, I_t$, and $R_t$ are latent. By solving the set of ODEs, we can discover the \emph{latent epidemic dynamics} which are described by the values of $\textbf{s}_{t}$ and $\Omega_{t}$ for the entire period for which we have observational data, i.e., $\forall t \in \{t_0,\ldots,t_N\}$.

\subsection{RNN architecture}
\label{subsec:rnn}
Our second building block is a Recurrent Neural Network (RNN) with attention, which has been extensively used in neural epidemic forecasting as a central building block ~\cite{adhikari_epideep:_2019,kamarthi_when_2021,wang_defsi_2019}.
Here, we introduce the base architecture of this model.
Informally, at prediction time $t=t_N$ we are given a multivariate time series of features/signals $\mathcal{X}=\left\{\textbf{x}_{t}\right\}_{t=t_0}^{t_N}$ with $\textbf{x}_{t}\in \mathbb{R}^{D_x}$, where $D_x$ is the number of features.
And we are tasked to predict $K$ steps ahead in the future. We encode the feature time series until $t_N$ by passing it through a Gated Recurrent Unit (GRU)~\cite{cho2014learning} to obtain a condensed representation for each time step: $\{\textbf{h}_t\}_{t=t_0}^{t_N} =  \text{GRU}(\left\{\textbf{x}_{t}\right\}_{t=t_0}^{t_N} )$
where $\textbf{h}_t$ is the hidden state of the GRU for time step $t$.
To capture long-term relations and prevent over-emphasis on last terms of sequence we use a self-attention layer~\cite{vaswani2017attention} which involves passing the embeddings into linear layers to extract meaningful similarities. 
Then, we use the attention weights to combine the latent representations and obtain a single embedding representing the time series of data features and pass it to a feedforward network to make the prediction $y_{t=N+k}$.

\subsection{PINNs for Systems Biology}
\label{subsec:pinn}
Recently, several works~\cite{yazdani_systems_2020,karniadakis_physics-informed_2021} in Systems Biology have used PINNs for solving forward and inverse problems with ODEs. The neural network $\text{N}(t)$ is a function of single variable $t$ and ODE system is in the form $f_{\text{ODE}}(t)$ describing the rate of change (gradient) of some function with respect to $t$. Gradient $\frac{d \text{N}(t)}{dt}$ can be  using computed via Automatic Differentiation--autograd, which in turn makes it possible to train the neural network $\text{N}(\cdot)$ while minimizing the residual between the two gradients, e.g. loss $\big(\frac{d \text{N}(t)}{dt} - f_{\text{ODE}}(t)\big)^2$.

\section{Problem formulation}
As mentioned earlier, we aim on harnessing the strengths of both machine learning/deep learning approaches (which have been very successful in short-term forecasting) and mechanistic models (which are useful for long-term trend projections). Hence, our problem is one of merging neural models with mechanistic model dynamics while maintaining benefits from both the techniques. 
To capture this specific intention, we modify traditional forecasting problems~\cite{adhikari_epideep:_2019} in the following manner:

\noindent\fbox{
    \parbox{0.95\columnwidth}{
    \textbf{\underline{Problem:}} \problem
\par\noindent\textbf{\underline{Given}}: $\bullet$ A base \textbf{epidemiological model} mathematically represented as a set of ODEs (for example, see the SEIRM and SIRS models in Section~\ref{subsec:seirm}). $\bullet$ A base \textbf{RNN} (See Section~\ref{subsec:rnn}). $\bullet$ \textbf{Data:} an observed multivariate time series of COVID/flu-related signals $\mathcal{X}=\left\{\textbf{x}_{t}\right\}_{t=t_0}^{t_N}$ and corresponding values for the forecasting target (new COVID-associated deaths or ILI flu counts) $\mathcal{Y}=\left\{y_{t}\right\}_{t=t_0}^{t_N}$, where $t_0$ is the first day of the outbreak and $t_N$ is the current date.
\par\noindent\textbf{\underline{Predict:}} next $K$ values of the forecasting target, i.e. $\left\{\hat{y}_{N+k}\right\}_{k=1}^{K}$ (here $K$ is the size of the forecasting window/horizon), such that predictions are \emph{accurate} and \emph{well-correlated} with the trends of the epidemic curve.
}
}

We are also interested in learning if taking advantage of selective superiorities of both approaches can push the prediction horizon (i.e., how many steps ahead we can forecast). 
Typically, CDC forecasting initiatives request short-term forecasts up to 4-weeks ahead ($K=4$) – see~\cite{cramer_evaluation_2021,jin_interseries_2021}. Longer forecasting horizons have not been explored much, thus, we propose the double of the current horizon (i.e., $K=8$) in this paper.

\section{Our Approach}

\begin{figure*}[t!]
    \centering
    \small
    \includegraphics[width=0.97\linewidth]{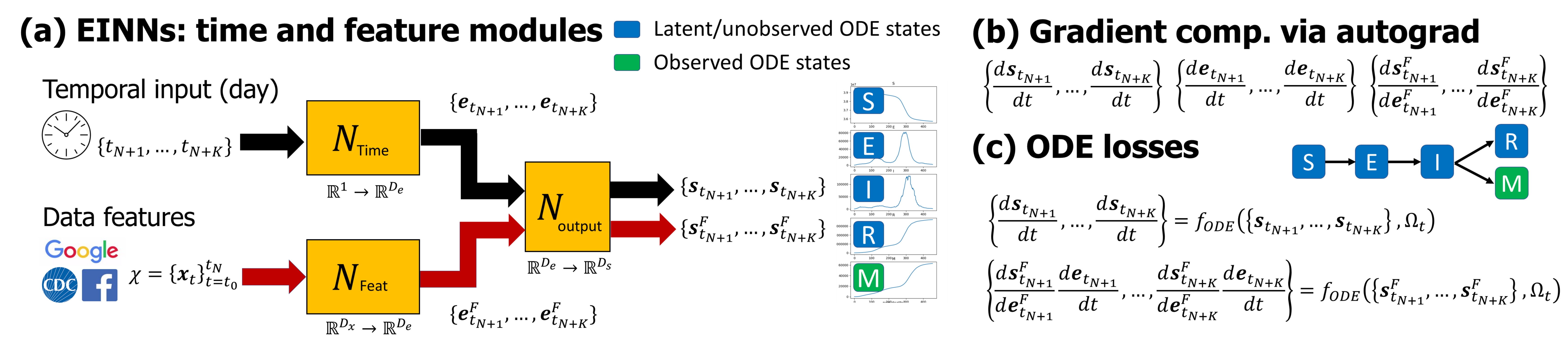}
    \caption{Depiction of our proposed framework \ourmethod. (a) The pipeline of EINNs has two independent computational paths: time module (source model) and feature module (target model). For simplicity, in our equations we refer to $N_{\text{Time}}$ as a single module, but, in practice, this is implemented as two neural models: $ N_{\text{output}} \circ N_{\text{Time}}$ where $N_{\text{output}}$ is a multi-layer feedforward network. Similarly, $N_{\text{Feat}}$ is implemented as $N_{\text{output}} \circ N_{\text{Feat}}$.
    During training step 2, when $\textbf{e}_{t} \approx \textbf{e}_{t}^{F}$, we will freeze layers and train only $N_{\text{output}}$.
    (b) Three gradients are computed via autograd:  $d\textbf{s}_{t}/dt$, $d\textbf{e}_{t}/dt$, and $d\textbf{s}_{t}^{F}/d\textbf{e}_{t}^{F}$. Using these gradients we can compute ODE loss for the time module and approximate  $d\textbf{s}_{t}^{F}/dt$ via our gradient trick.
    (c) We utilize equations $f_{\text{ODE}}$ to compute the ODE losses.
    Approximation of $d\textbf{s}_t^{F}/dt$ is used to compute the ODE loss for the feature module $\mathcal{L}^{\text{ODE-F}}$. This loss encourages integration of ODE dynamics from the time module (source model) into the feature module (target model).
    }
    \label{fig:method}
\end{figure*}

To tackle the problem \problem, one can easily conjure `na\"ive' approaches. A simple approach is to calibrate the given mechanistic model with the observed targets $\mathcal{Y}$, and train the base RNN using the generated (synthetic) curve. Similarly, one could also use the ODEs to regularize the neural predictions or could train an ensemble with neural network's and ODE-model's predictions. However, as we show later in our experiments, while these approaches often can maintain the performance of the base RNN, they do not generate well-correlated predictions.

\par\noindent{\textbf{\underline{Overview } }}
See Figure~\ref{fig:method}.
We propose using an heterogeneous domain transfer learning setup~\cite{moon2017completely}, where we transfer knowledge from a source model to a target model. Here our source model is a PINN whose purpose is discovering the latent epidemic dynamics (solving ODEs). The gradients of our ODE epi-models ($d\textbf{s}_t/dt$) are with respect to time; therefore, as noted in Section~\ref{subsec:pinn}, time is the only input to this PINN. Thus, we refer to this PINN as \underline{\emph{time module}}. 
The target model is an RNN which ingests data features from heterogeneous sources--thus we call it \underline{\emph{feature module}}--and incorporates appropriate inductive bias to model sequential time series data.
Note that both source and target models predict the same output $\textbf{s}_t$, which are the ODE states.
Therefore, the feature module learns a mapping from a multivariate time series of COVID/flu-related signals $\mathcal{X}$ (exogenous data features, i.e., not-represented in the ODE) to the epidemic dynamics $\textbf{s}_t$.
Next, we explain each of these modules in detail.

\subsection{Time module (source model): learning latent time-varying dynamics}
\label{subsec:time}
The time module interfaces with the set of ODEs describing an epidemic mechanistic model. Via PINNs, it learns the latent epidemic dynamics given observational data.
Following the introduction in Section~\ref{subsec:pinn}, PINNs solve the ODEs by jointly minimizing the observational error and the residual between gradient given by the ODE and the gradient of the neural network with respect to the time inputs (computed via autograd). As in most literature--see Section~\ref{sec:related_work}--our time module $N_{\text{\tiny Time}}(t)$ is parametrized by a multi-layer perceptron that ingests time $t$ as input and predicts ODE states for time $t$, denoted as $\textbf{s}_t \in \mathbb{R}^{D_s}$, where $D_s$ is the number of ODE states (e.g., 5 for SEIRM). 
We want this neural network to make predictions that follow epi dynamics described the set ODEs $f_{\text{\tiny ODE}}$. This is, we make $N_{\text{\tiny Time}} (t) = \textbf{s}_t \text{; subject to } d\textbf{s}_t/dt =  f_{\text{\tiny ODE}}(\textbf{s}_{t}, \Omega_t),$
where $\Omega_t$ are the learned ODE parameters for time $t$.
We minimize the ODE loss (unsupervised loss $\mathcal{L}^{\text{ODE-T}}$) while fitting the observed data (supervised loss $\mathcal{L}^{\text{Data-T}}$):
\begin{gather}
    \mathcal{L}^{\text{ODE-T}} = \frac{1}{N+1} \sum_{t=t_0}^{t_N} \left[ \frac{d\textbf{s}_t}{dt} - f_{\text{\tiny ODE}}(\textbf{s}_{t}, \Omega_t) \right]^{2} \\
    \mathcal{L}^{\text{Data-T}} = \frac{1}{N+1} \sum_{t=t_0}^{t_N} \left[ \hat{M}_{t} - M_t \right]^{2} 
\end{gather} 
where $\hat{M}_t$ is predicted mortality by the time module.
To discover the latent dynamics, we want to obtain $\textbf{s}_{t}$ and $\Omega_t$.

\par\noindent{\textbf{\underline{Constraining the optimization via epi domain knowledge} }} 
In contrast to the setting in most prior work in PINNs where most states of the system dynamics are observed, most states in our \ode model are latent. We found learning in such a scenario can be very challenging. We alleviate this by infusing additional epidemiological domain knowledge in the form of monotonicity constraints.
In particular, we adapt monotonicity losses from \cite{muralidhar_incorporating_2018} to our setting. They proposed to penalize consecutive predictions  if they are not monotonic in the required direction. Note that the difference between consecutive predictions are discrete approximation to the derivatives. Here, we generalize this loss to continuous derivatives by taking the limit $\lim_{t \to 0}$.
Now, derivatives can be directly computed via autograd.  
To incorporate these constraints, we use domain knowledge. Using SEIRM as an example (similar can be easily derived for other epidemiological models), we know that the Susceptible state $S_t$ monotonically decreases and the Recovered state $R_t$ monotonically increases. Then, we add a penalty when $dS_t/dt$ is positive and when $dR_t/dt$ is negative as follows: 
\begin{multline}
    \mathcal{L}^{\text{Mono}} = \frac{1}{N+1} \Big( \sum_{t=t_0}^{t_N} \left[ \frac{dS_t}{dt} \text{ReLU}(\frac{dS_t}{dt}) \right]^{2}
    + \\
    \sum_{t=t_0}^{t_N} \left[ -1 \frac{dR_t}{dt} \text{ReLU}(-\frac{dR_t}{dt}) \right]^{2} \Big),
\end{multline}
where $\text{ReLU(x)}=\max(0,x)$ is the rectified linear function. Note that $S_t$ and $R_t$ are part of $\textbf{s}_t$, which is the output of the time module; thus, $\frac{dS_t}{dt}$ and $\frac{dR_t}{dt}$ are computed via autograd.

\par\noindent{\textbf{\underline{Coping with spectral bias in neural networks}}}
One of the central issues in fitting PINNs is the spectral bias of neural networks, which is the tendency of neural networks to fit low frequency signals ~\cite{wang_understanding_2021}. 
To overcome this, usually the  neural networks are given more flexibility to fit high frequency systems.
Here, we adopted Gaussian Random Fourier feature mappings ~\cite{tancik_fourier_2020}: $\Gamma(v) = [\cos(2\pi \textbf{B} \textbf{v}), \sin(2\pi \textbf{B} \textbf{v})]^T $, where each entry in $\textbf{B} \in \mathbb{R}^{d \times 1}$ is sampled from $\mathcal{N}(0,\sigma^2)$, where $\sigma$ is a hyperparameter.

\par\noindent{\textbf{\underline{Handling time-varying ODE parameters}}}
As mentioned earlier, our ODE model is time varying, therefore we have to learn mechanistic parameters for each time step, which increases the difficulty of the optimization. To make this more tractable, we propose a consistency loss between consecutive parameters.
\begin{equation}
    \mathcal{L}^{\text{Param}} = \frac{1}{N+1} \sum_{t=t_0}^{t_N} \left[ \Omega_{t+1} - \Omega_{t} \right]^{2}
\end{equation}

\subsection{Feature module (target model): connecting features to epidemic dynamics via gradient matching}
\label{subsec:feature}
The feature module is composed of a based RNN and ingests multivariate time-series of COVID/flu-related signals $\mathcal{X}$ to predict the epidemic dynamics $\textbf{s}^F_{t}$. (Note: embeddings and outputs of feature module have superscript $F$ to avoid confusions with the ones from the time module). 
Here, we want to ensure that the predictions made by the feature module are consistent with the ones given by the ODE model. Hence, we want the feature module neural network $N_{\text{Feat}}(\mathcal{X})$ to follow the learned latent dynamics from the time module neural network $N_{\text{Time}}(t)$ (note the heterogeneous domains). 
We can formalize our goal as follows:
$N_{\text{Feat}}(\mathcal{X}) = \textbf{s}^F_t \text{; subject to } \frac{d\textbf{s}^F_t}{dt} =  f_{\text{\tiny ODE}}(\textbf{s}^F_{t}, \Omega_t)$, where $\textbf{s}^F_t \in \mathbb{R}^{D_s}$, are the ODE states predicted by the feature module  and $\Omega_t$ are the same ODE parameters used by in the time module. 

\par\noindent{\textbf{\underline{Matching the ODE gradient}}} 
We cannot directly calculate $\frac{d\textbf{s}^F_t}{dt}$ via autograd from the inputs as we did for the time module because our base \rnn ingests features. We propose to use internal representations (embeddings) so that we can approximate the gradient to an expression that can be computed via autograd directly. 
Let $\textbf{e}_{t}\in \mathbb{R}^{D_e}$ and $\textbf{e}_{t}^{F}\in \mathbb{R}^{D_e}$ be embeddings for the time module and feature module, respectively ($D_e$ is embedding size). Then, by using the chain rule, we propose to approximate the gradient of $\textbf{s}^F_t$ assuming $\textbf{e}_{t} \approx \textbf{e}_{t}^{F}$ and have our \emph{gradient trick}:
\begin{equation}
    \frac{d\textbf{s}_t^{F}}{dt} = \frac{d\textbf{s}_{t}^{F}}{d\textbf{e}_{t}^{F}} \frac{d\textbf{e}_{t}^{F}}{dt} \approx \frac{d\textbf{s}_{t}^{F}}{d\textbf{e}_{t}^{F}} \frac{d\textbf{e}_{t}}{dt}
    \label{eq:gradapprox}
\end{equation}
where $\frac{d\textbf{s}_{t}^{F}}{d\textbf{e}_{t}^{F}}$ can be calculated in the feature module using autograd because $\textbf{e}_{t}^{F}$ is the only variable that is needed to compute $\textbf{s}_{t}^{F}$. Similarly, $t$ is the only input needed for computing $\textbf{e}_{t}$, thus, we can use autograd to compute $\frac{d\textbf{e}_{t}}{dt}$.
To make this approximation valid, we have to make these embeddings $\textbf{e}_{t}$ and $\textbf{e}_{t}^{F}$ similar. We do this with the following loss:
\begin{equation}
    \mathcal{L}^{\text{Emb}} = \frac{1}{N+1} \sum_{t=t_0}^{t_N} \left[ \textbf{e}_{t} - \textbf{e}_{t}^{F} \right]^{2} 
    \label{eq:kd_emb}
\end{equation}
This derivation allows us to make the feature module to learn the gradients learned by the time module by minimizing an ODE loss for the feature module: 
\begin{equation}
    \mathcal{L}^{\text{ODE-F}} = \frac{1}{N+1} \sum_{t=t_0}^{t_N} \left[ \frac{d\textbf{s}_{t}^{F}}{d\textbf{e}_{t}^{F}} \frac{d\textbf{e}_{t}}{dt} - f_{\text{\tiny ODE}}(\textbf{s}^{F}_{t}, \Omega_t) \right]^{2} 
    \label{eq:gradmatching}
\end{equation}

\par\noindent{\textbf{\underline{Aligning with data and time module outputs}}}
Matching the ODE gradient is not enough to ensure the dynamics will be transferred. We have to make sure that the feature module outputs are aligned with data and with the ODE dynamics as found by the time module.
For fitting the data, we define data loss in a manner similar to the time module:
\begin{equation}
    \mathcal{L}^{\text{Data-F}} = \frac{1}{N+1} \sum_{t=t_0}^{t_N} \left[ \hat{M}^{F}_{t} - M_t \right]^{2}
\end{equation}
where $\hat{M}^{F}_{t}$ is the predicted mortality
of the feature module. To align the time and feature modules, we use knowledge distillation (KD)~\cite{ba2014deep}, a popular transfer learning method. We impose our KD loss on the outputs of these two:
\begin{equation}
    \mathcal{L}^{\text{Output}} = \frac{1}{N+1} \sum_{t=t_0}^{t_N} \left[ \textbf{s}_{t} - \textbf{s}_{t}^{F} \right]^{2}. 
\end{equation}

Note that our time module is able to predict for any given time but our base \rnn makes prediction for one target in the future. To align these two, we make our feature module to make joint prediction using a decoder GRU which takes $\textbf{u}_{t_0:t_N}$ as the initial hidden state and roll the GRU forward for every prediction step ahead taking time as input. Thus, our decoder equations will be
    $\textbf{e}_{t=t_{N+k}} =  \text{GRU}( \textbf{u}_{t_0:t_N}, t_{N+k})$ and our final predictions $y_{t=t_{N+k}} =  \text{FFN}( \textbf{e}_{t=t_{N+k}})$.

\subsection{Model training, inference, and implementation}
\par\noindent{\textbf{\underline{Training}}} $\bullet$ Step 1: 
During the first training step, our goal is to make $\textbf{e}_{t} \approx \textbf{e}_{t}^{F}$ so that later we can use the gradient approximation stated in Equation~\eqref{eq:gradapprox} so that we can then match the gradient of the ODE. 
For this, we can train all parameters (neural networks and ODE parameters $\Omega_t$) jointly with all the losses except for the feature ODE loss $\mathcal{L}^{ODE_F}$. 
$\bullet$ Step 2: Once $\mathcal{L}^{\text{Emb}}$ is small, we can train all losses together, however, it might be unstable and  $\mathcal{L}^{\text{Emb}}$ may start to increasing which in turn invalidates our gradient matching trick and makes the minimization of $\mathcal{L}^{\text{ODE}_F}$ misleading. We found the training is more stable when freezing all previous layers to $\textbf{e}_{t}$ in the time module and all previous layers to $\textbf{e}_{t}^{F}$ in the feature module. In this case, we only focus the learning in the last layers, therefore, they should have enough representation power for this task.
\textbf{\underline{Inference}} 
At inference, although we have predictions from both time and feature modules, we  solely utilize the feature module predictions as it ingests features and we want to emphasize the utility of inserting dynamics in ML models. 
\textbf{\underline{Implementation}} As previous work notes~\cite{yazdani_systems_2020}
learning ODE dynamics with PINNs is challenging. We found it useful to bound ODE parameters $\Omega_t$ and initialize them based on the numerical solution of the ODE as output by a standard solver, e.g., Nelder-Mead. We keep $\Omega_t$ kept learnable through the training process because the neural networks may find a better solution.
More details are in our appendix. 

\section{Experiments}
\label{sec:expt}

\subsection{Setup, metrics, and baselines}
\label{subsec:setup}
\begin{table*}[t!]
\centering
\caption{
\ourmethod is the only one consistently providing accurate and well-calibrated forecasts and it is among the best performing for all metrics (lower NR1, NR2 and ND is better; higher Pearson correlation is better). Top 2 models per column are in bold (including tied models).
These results are an average across 5696 predictions per model. 
} 
\resizebox{0.70\textwidth}{!}{
\begin{tabular}{@{\extracolsep{8pt}}lccccccc}
\toprule

{} & \multicolumn{3}{c}{}  & \multicolumn{3}{c}{} & {\textbf{Trend}}\\
{} & \multicolumn{3}{c}{\textbf{Short-term (1-4 wks)}}  & \multicolumn{3}{c}{\textbf{Long-term (5-8 wks)}} & {\textbf{correlation}}\\
 \cmidrule{2-4} 
 \cmidrule{5-7} 
 \cmidrule{8-8} 
 \textbf{Model} & \textbf{NR1} & \textbf{NR2} & \textbf{ND} & \textbf{NR1} & \textbf{NR2} & \textbf{ND} & \textbf{PC} \\ 
 \midrule \midrule
 \multicolumn{8}{c}{\textbf{Task 1: COVID-19 Forecasting (US National + 47 states)}} \\ 
\midrule \midrule
\rnn (GRU+Atten) & 
1.09 &	0.50 &	0.86 &	1.19 &	0.53 &	0.96 & 0.08 \\
Mechanistic model (\ode) & 2.35 &	1.13 &	1.36 &	7.14 &	2.99 &	3.11 &	\textbf{0.53} \\	
\midrule
\generation  & 
0.79 &	0.35 &	0.60 &	\textbf{0.93} &\textbf{	0.40} &	0.74 & -0.01 \\
\regularization  &
1.05 &	0.48 &	0.81 &	1.19 &	0.53 &	0.97 &	0.09 \\	
\ensembling  & 
0.91 &	0.41 &	0.68 &	\textbf{0.93} &	\textbf{0.40} &	\textbf{0.69} & -0.01 \\ 
\textbf{\ourmethod (ours)}  & \textbf{0.54}	& \textbf{0.24} &	\textbf{0.38} &	\textbf{0.85} &	\textbf{0.37} &	\textbf{0.66} &	\textbf{0.46} \\ \midrule
\pinn (time module standalone) & 
0.84 &	0.38 &	0.64 &	\textbf{0.93} &	\textbf{0.40} &	0.72 &	0.24 \\
\ourmethod-NoGradMatching  & \textbf{0.64} &	\textbf{0.29} &	\textbf{0.49} &	0.98 &	0.43 &	0.79 &	0.03 \\
\midrule \midrule
\multicolumn{8}{c}{\textbf{Task 2: Influenza Forecasting (10 HHS regions)}} \\ \midrule \midrule
\rnn (GRU+Atten) & 
0.72 &	0.38 &	0.67 &	1.19 &	0.51 &	1.14 &	-0.03\\
Mechanistic model (\odeflu) & 
0.72 &	0.38 &	0.51 &	1.16 &	0.55 &	0.81 &	\textbf{0.71} \\ \midrule
\generation  & 
0.76 &	0.4 &	0.71 &	1.21 &	0.52 &	1.15 &	-0.14 \\
\regularization  & 
1.19 &	0.64 &	1.00 &	1.22 &	0.54 &	0.9 & -0.45 \\
\ensembling  & 
0.89 &	0.47 &	0.77 &	\textbf{0.83} &	\textbf{0.35} &	\textbf{0.73} &	-0.69 \\
\textbf{\ourmethod (ours)}  & \textbf{0.53} & \textbf{0.27} &	\textbf{0.37} &	\textbf{1.01} &	\textbf{0.42} &	\textbf{0.73} &	\textbf{0.68} \\ \midrule
\pinn (time module standalone) &
0.55 &	0.29 &	0.44 &	1.13 &	0.48 &	1.02 &	-0.47 \\
\ourmethod-NoGradMatching  & \textbf{0.53} &	\textbf{0.27} &	\textbf{0.38} &	1.02 &	\textbf{0.42} &	0.76 &	0.50 \\
\bottomrule
\end{tabular}
}
\label{tab:results}
\vspace{-10pt}
\end{table*}
All experiments were conducted using a 4 Xeon E7-4850 CPU with 512GB of 1066 Mhz main memory and 4 GPUs Tesla V100 DGXS 32GB. Our method implemented in PyTorch (implementation details in the appendix) trains on a GPU in about 30 mins for one predictive task. Inference takes only a few seconds. 
Appendix, code, and other resources can be
found online\footnote{Resources website: \url{https://github.com/AdityaLab/EINNs}}.

\par\noindent{\textbf{\underline{Evaluation}}}
All the results are for forecasting COVID-19 mortality in the US up to 8-weeks ahead in the future.  
For COVID-19, as per previous CDC-coauthored papers~\cite{cramer_evaluation_2021}, we evaluate at state and national level. Specifically, we include 47 states; we exclude 3 out of the 50 states where the \ode mechanistic model had difficulties fitting due to sparsity of death counts (specifically Alaska, Montana, and Wyoming). 
Our evaluation period is 8 months 
from Sept. 2020 to March 2021. 
which includes the complete Delta wave in the US, and we make 8-weeks ahead predictions for every two weeks in this period. We used June 2020 to Aug. 2020 to tune our models.
For flu, we also follow CDC~\cite{biggerstaff_results_2018} and predict for all 10 HHS region\footnote{\url{hhs.gov/about/agencies/iea/regional-offices/index.html}} for a period of 5 months (Dec. 2017 to May 2018).
For each forecasting week, all models are trained with historical data available until that week (i.e., they are trained on every prediction week). 
In total, we make 5696 predictions per model which requires training each of them 700+ times, therefore, is very computationally expensive to run all models for multiple runs.

\par\noindent{\underline{\textbf{Metrics}}}
Our focus is in predictions that are accurate and well-correlated the epidemic trends, thus we measure two aspects of our predictive performance: error and trend correlation. 
$\bullet$ \textbf{Error metrics}: 
As previous work in this area~\cite{adhikari_epideep:_2019}, we adopt metrics based on root mean squared error and absolute deviation. 
Because the number of deaths largely vary across regions, we use normalize versions of popular error metrics so that we can aggregate performance over multiple regions.
We use two different versions of Normalized Root Mean Squared Error (NR1 and NR2) and Normal Deviation (ND) following \cite{roy2021deep,remy2021overcoming}. These metrics are described in detail in our appendix.
For all of these metrics, we calculate them at short-term forecasting (1-4 weeks) and long-term forecasting (5-8 weeks) and calculate their mean value.
$\bullet$ \textbf{Correlation metric}: Following~\cite{deng_cola-gnn_2020}, we use Pearson correlation and use their median across weeks.

\par\noindent{\underline{\textbf{Data}}}
We collected important publicly available signals from a variety of trusted sources that are relevant to \covid forecasting. For COVID-19, we collected 13 features in this dataset, this include mobility from Google 
and Apple
, social media surveys from Facebook,  
hospitalization data from the U.S. Depart. of Health \& Human Services  
and CDC
, and cases and mortality from Johns Hopkins Univ.
For flu, we use the 14 signals from the Google symptom dataset. See appendix for more details and links.

\par\noindent\textbf{\underline{Baselines}}
As we are the first to pose the \problem problem, we do not have off-the-shelf baselines. Instead, our focus is on how to incorporate ODE dynamics into NNs. Hence we focus on the different ways these have been explored in literature~\cite{dash2022review}. 
$\bullet$ \generation: Similar to~\cite{wang_defsi_2019,sanchez2020learning}, the NN learns directly from data generated by the numerical solution of  \ode/SIRS. 
$\bullet$ \regularization: Similar to~\cite{gao_stan_2021,gaw2019integration}, the NN predicts both the ODE states and the ODE parameters. Then uses the ODE parameters to regularize the states via a loss based on the \ode/SIRS equations. 
$\bullet$ \ensembling: As per \cite{adiga_all_2021,yamana2017individual},  combines predictions from \rnn and \ode/SIRS via a NN that outputs final predictions.

\subsection{Results in COVID-19 and influenza}
Our results showcase EINN as an effective general framework for incorporating epidemic dynamics from a mechanistic model into a neural network. We first demonstrate that we can leverage advantages from both modeling paradigms resulting in consistently good forecasts across all tasks and metrics. We also compare against other non-trivial methods to incorporate ODE dynamics into neural models. 
To contextualize our model's performance with the broader picture of epidemic forecasting, we also have results with standard data-driven baselines, which can be found in our appendix.

\par\noindent{\textbf{\underline{Q1:} Leveraging advantages of both mechanistic models and neural models.}}
Our \rnn has a lower or similar error in short- and long-term forecasting than the \ode and SIRS, but its predictions are much less correlated with epidemic trends (see lines 1-2 of comprehensive results in Table~\ref{tab:results}). 
By integrating mechanistic and neural models, \ourmethod is capable of taking advantage of both. 
Comparing \ourmethod with the SEIRM/SIRS, Pearson correlations are close but our predictions are much more accurate (up to 77\% less error). 
Indeed, \ourmethod 
not only improves \rnn correlation by 475\% but also its accuracy up to 55\% thanks to the incorporation of short and long-term dynamics.
Note that our goal was not to beat the SEIRM/SIRS but have a method that has a consistently good performance across accuracy and correlation.

\par\noindent{\textbf{\underline{Q2:} Benefits over other ways to incorporate epidemic dynamics into neural models.}}
\ourmethod has the lowest error and best correlation in comparison with other existing ways to incorporate epidemic dynamics to neural networks.
We can see that these methods may excel in one task (e.g., \ensembling in long-term forecasting) but they are severely worse in other important tasks. Instead, \ourmethod is the only one consistently good in all tasks. 

\par\noindent{\textbf{\underline{Q3:} Ablation: time module PINN and gradient matching.}}
We perform ablation studies to understand what are the contributions of the main components of our model. First, we analyze our time module trained standalone, i.e., being trained without the feature module with losses in Equations (5-9) (\pinn in Table~\ref{tab:results}). 
We can see that, although our time module \pinn directly interacts with the ODE and their behavior will be coupled during training, they have different behavior in test. In fact, this points to the need that we need features to be able to extract representations that generalize in test.
Second, we assess the contribution of our gradient matching trick (\ourmethod-NoGradMatching), for which we train with all losses except for the ones in Equations~\eqref{eq:kd_emb} and \eqref{eq:gradmatching}. In this scenario where only $\mathcal{L}^{\text{Output}}$ helps to transfer the dynamics, we can see that it is a less effective way.

\par\noindent{\textbf{\underline{Q4:} Case-study} and \textbf{\underline{Q5:} Sensitivity to hyperpameters.}}
In our appendix, we conducted a case study in US National and New York to visually analyze the advantages of our method in both accuracy and correlation (see example for US National in  
In Figure~\ref{fig:teaser}b).  
We found most hyperparameters are not sensitive. See appendix for more results and details.

\section{Discussion and Societal Impact}
\label{sec:conclusion}

The COVID-19 pandemic has impacted possibly every aspect of life and has exemplified our vulnerability to major public health threats. 
This underscores the importance of infectious disease detection and prediction for shaping more resilient societies.
Preventing and responding to such pandemics requires trustworthy epidemic forecasts, e.g. forecasts well correlated with actual epidemic trends.

The ML community has been very active in CDC forecasting initiatives and has harnessed multiple successes. However, generating trustworthy epidemic forecasts may require more than only data. In this paper we tackle this challenge by introducing \ourmethod to incorporate mechanistic dynamics (via the SEIRM/SIRS model) into neural models (using a RNN style base model).  We show the effectiveness of a principled method to transfer relevant knowledge via gradient matching of the ODE equations, without integrating (forward pass) the ODE model. 
Through extensive experiments over states/regions in the US we also show the usefulness of \ourmethod in COVID-19 and flu forecasting and also the importance of our various design choices. Overall, we believe this work opens up new avenues for leveraging epi domain knowledge into AI models for better decision making. Connecting complex mechanistic models to neural networks also enables us to have learned representations useful for other tasks downstream like what-if predictions, which would be worth exploring. In addition investigating more complex epidemiological models (like network based agent models) would be fruitful.

\section*{Acknowledgments}
This work was supported in part by the NSF (Expeditions CCF-1918770, CAREER IIS-2028586, RAPID IIS-2027862, Medium IIS-1955883, Medium IIS-2106961, CCF-2115126), CDC MInD program, ORNL, faculty research award from Facebook and funds/computing resources from Georgia Tech. B.A. was supported by CDC-MIND U01CK000594 and start-up funds from University of Iowa. NR was supported by US NSF grants Expeditions CCF-1918770, NRT DGE-1545362, and OAC-1835660. We also would like to thank Harsha Kamarthi 
for his helpful suggestions which improved the paper.

\clearpage
\normalsize
\appendix

\section{Background (continuation)}
\noindent\par\textbf{SIRS model for seasonal influenza.} The SIRS model has been extensively used for modelling the seasonal influenza outbreaks. Here we use the version proposed in~\cite{shaman2010absolute} which other follow up papers have used to guide their analysis (e.g., ~\cite{pei2020aggregating}). The model consists of three compartments: Susceptible ($S_t$), Infected ($I_t$), and Recovered ($R_t$). It is parameterized by three variables $ \Omega = \{ \beta, D, L\}$, where $\beta$ is the infectivity rate, $D$ is the mean duration of immunity, and $L$ is the mean duration of the immunity period. Note that $N-S-I$ is the number of immune individuals. 
As per in SEIRM, we utilize time-varying parameters 
$\Omega_{t} = \{ \beta_t, D_t,  L_t\}$ 
at the given time-stamp $t$ and denote the ODE states as $\textbf{s}_{t} = [ S_t, I_t, R_t]^T$.
The ODEs describing the model is as follows:
\begin{gather}
\frac{d S_t}{d t}=\frac{N-S_t-I_t}{L_t}-\frac{\beta_t I_t S_t}{N} \\\nonumber
\frac{d I_t}{d t}=\frac{\beta_t I_t S_t}{N}-\frac{I_t}{D_t}
\end{gather}

\noindent\par\textbf{Calibration. } 
The target/calibration variable for COVID-19 is associated mortality, while in flu it is percentage of patients with influenza-like-illness (ILI) symptoms, which is not directly represented in the SIRS model. We provide details in Section~\ref{sec:eval_appendix} on how to connect ILI to the SIRS model.

\section{More details on data}
Our datasets were collected from publicly available sources\footnote{Data links:
\url{apple.com/covid19/mobility}; \url{google.com/covid19/mobility}; \url{coronavirus.jhu.edu}; \url{healthdata.gov}; \url{delphi.cmu.edu}; \url{gis.cdc.gov/grasp/COVIDNet/COVID19_3.html};
\url{goo.gle/covid19symptomdataset}
}. We describe them in details as follows.
\begin{itemize}
    \item Data signals 1: Mobility signals. The signals originate from the record of people visiting points of interest (POI) in various regions. According to Google, daily changes in visits to various POI categories are collected and compared with the period January 3 - February 6, 2020. Additionally, we collected a daily change of visitors from Apple, which shows the relative volume of directions requested across different US states compared to January 13. Different non-pharmaceutical interventions (NPIs) and different policies adopted by different states are implicitly illustrated by mobility signals.
    \item Data signals 2: Symptomatic surveys. Every day, Facebook collects statistics on COVID-like illness (\%CLI) and influenza-like illness (\%ILI) across the US and different states. On the basis of symptoms reported in voluntary surveys, they estimate this percentage.
    \item Data signals 3: Symptom search data. Google collects records of searches related to symptoms for multiple conditions and syndromes across the US and different states. Their system provides a metric that quantifies search volume associated with specific symptoms, which undergoes a privacy-protecting mechanism before being publicized. There are 400+ symptoms available in this dataset dating back to 2017, from which we only use a subset of 14. These are the following which are symptoms associated with influenza: Fever, Low-grade fever, Cough, Sore throat, Headache, Fatigue, Vomiting, Diarrhea, Shortness of breath, Chest pain, Dizziness, Confusion, Generalized tonic–clonic seizure, and Weakness.
    \item Data signals 4: Number of hospitalizations. The US Department of Health \& Human Services provides daily hospitalization admissions dating back to January 1, 2020. Several primary sources provide facility-level granularity reports to create this signal: (1) HHS TeleTracking, (2) reporting provided to HHS Protect by state/territorial health departments on behalf of their healthcare facilities, and (3) the National Healthcare Safety Network.
    \item Data signals 5: Number of new deaths. The Johns Hopkins University reports daily mortality for COVID-19. They collect and curate data from official websites of state public health departments across the US. This has been the source of data for the CDC COVID-19 forecasting initiative (Craemer et al. 2022).
    \item Data signals 6: weighted Influenza-like Illness (wILI). Time series data are collected by CDC from over 3,500 outpatient healthcare providers in the Outpatient Influenza-like Illness Surveillance Network (ILINet). Health care providers report voluntarily every week the percentage of patients with Influenza-like Illness (ILI) symptoms. ILI is defined as “fever (temperature of 100◦F [37.8◦C] or greater) and a cough and/or a sore throat without a known cause other than influenza.” This has been the source of data for previous iterations of the CDC FluSight forecasting initiative (Reich et al. 2019).
\end{itemize}

\section{Experimental setup (extra details)}

Code and data are available attached as a supplement.  
They both will be made public upon acceptance.

\par\noindent\textbf{Computational setup.}
All experiments were conducted using a 4 Xeon E7-4850 CPU with 512GB of 1066 Mhz main memory and 4 GPUs Tesla V100 DGXS 32GB. Our method implemented in PyTorch (implementation details in the appendix) trains on GPU in about 30 mins for one predictive task. Inference is takes only a few seconds.

\par\noindent\textbf{Real-time forecasting.}
We follow the literature on evaluating epidemic forecasting methodologies~\cite{shaman_real-time_2013,kamarthi_camul_2021,adhikari_epideep:_2019} and use the \emph{real-time forecasting} setup. 
We simulate real-time forecasting by making models train \emph{only} using data available until each of the prediction weeks and make predictions for 1 to 8 weeks ahead in the future. Data revisions in public health data are large and may affect evaluation and conclusions~\cite{kamarthi2021back2future,cramer_evaluation_2021}, therefore, we utilize fully revised data following previous papers on methodological advances~\cite{adhikari_epideep:_2019,rodriguez_steering_2021}.

\par\noindent\textbf{Evaluation details. }
Some models may make daily predictions while others weekly predictions. We follow CDC evaluation papers~\cite{cramer_evaluation_2021,reich_collaborative_2019} and convert all forecasts to weekly.
For this, we sum over 7 days for COVID-19 and take a 7-day average for flu.

\section{Implementation details}
\subsection{Data Preprocessing} 
\par\noindent\textbf{Feature scaling.}
Time-series of exogenous features can have wide range of values (e.g., number of confirmed cases vs percentage of people with COVID-like symptoms in social media surveys)
Therefore, we scale all signals per each region for which we use standard scaling (normalization).

\par\noindent\textbf{Time series for training.}
Although we may have long time series, we found our RNN works better with no chunking. As during training we have variable-length input sequences, we use a mask that is utilized when calculating the attention scores. As we follow the real-time forecasting setup, at inference time we use the complete input sequence thus we do not need a mask.

\subsection{Architecture and hyperparameters}
We describe in detail the hyperparameters for \ourmethod used for our experiments.
As mentioned in Section 5, we used data from June 2020 to Aug. 2020 for model design and hyperparameter tuning. 

\par\noindent$\bullet$ \textbf{Time module:} It is a feedforward network with input layer layers 40x40x40x20 followed by output layers 20x40x40x5, and activation function $\tanh$. 
Note that the input to this module is time, which is of dimension 1, but this is immediately transformed to 40 different signals via Gaussian Random Fourier feature mapping which then enter to the neural network.
Between the input and output layers, we have our embeddings $\textbf{e}_{t}$, which are of smaller size (20) than the other hidden layers. We make this selection because we want to make embeddings $\textbf{e}_{t}$ and $\textbf{e}_{t}^{F}$ to be as close as possible, and this is hard to achieve when we deal with high-dimensional embeddings as it has been noted in the Contrastive Learning literature~\cite{liu2021self}. Passing $\textbf{e}_{t}$ to the output layer gives us $\textbf{s}_{t}$.
With respect to the learnable ODE parameters, we a $\tanh$ transformation following~\cite{yazdani_systems_2020}. This ensures that the actual parameter values will be within their corresponding domain (0-1). In our experience, we found that initializing the ODE parameters with the ones found by the analytical solution works best.

\par\noindent$\bullet$ \textbf{Feature module:} As encoder, we used a bi-directional GRU with 2 layers and hidden states of dimension 32. As decoder, we use another bi-directional GRU with 1 layer and hidden states of dimension 32 and an feedforward output layer of 32x20 to obtain $\textbf{e}_{t}^{F}$. To encourage transfer learning, we utilize shared layers between the time and feature modules. Therefore, the embedding $\textbf{e}_{t}^{F}$ is passed to the output layer of the time module to obtain $\textbf{s}_{t}^{F}$.

\par\noindent$\bullet$ \textbf{Loss weights:} All our results use loss weight of 1 except for the following losses. For our ODE loss $\mathcal{L}^{\text{ODE}}$  $\mathcal{L}^{\text{ODE}-F}$ we use weight loss of 10, and we weight the same in our monotonicity loss $\mathcal{L}^{\text{Mono}}$.
Our parameter consistency loss $\mathcal{L}^{\text{Param}}$ is weighted with $0.001$. Finally, we have a helper loss for the time module that ingests data from the analytical solution of the ODE, to which we put a weight of $0.1$

\section{Evaluation metrics}
\label{sec:eval_appendix}
As noted in Section 5, we used two different versions of Normalized Root Mean Squared Error (NR1 and NR2) and Normal Deviation (ND) following~\cite{roy2021deep,remy2021overcoming}. Given prediction $\hat{y}_{w,k}$ at prediction week $w$ for $k$-weeks ahead in the future, and the corresponding ground truth value $y_{w,k}$, our error metrics are the following:
\begin{gather*}
\text{NR1} = \frac{\sqrt{\frac{1}{W K} \sum_{w,k} (y_{w,k} - \hat{y}_{w,k})^2}}{\frac{1}{N K} \sum_{w,k}|y_{w,k}|} \\ 
\text{NR2} = \frac{\sqrt{\frac{1}{W K} \sum_{w,k} (y_{w,k} - \hat{y}_{w,k})^2}}{ \max(y_{w,k}) - \min(y_{w,k}) } \\
\text{ND} = \frac{\sum_{w,k}|y_{w,k} - \hat{y}_{w,k}|}{\sum_{w,k}|y_{w,k}|} 
\end{gather*}
where $W$ is the number of predictions weeks and $K$ is number of weeks ahead in the future; for our setup we have $W = 14$ and $K=8$, which makes 112 different predictions for model for a single region and makes 5376 predictions per model over all regions. 
Note: In the case of COVID-19, there are several states for which there was no death in a particular week, therefore, we add 1 death to the denominator to avoid numerical issues. In flu, we do not need this.

Regarding correlation, we use Pearson correlation over the sequence of 8 predictions in the future as per~\cite{deng_cola-gnn_2020}:
\begin{gather*}
    \text{PC}=\frac{\sum_k (y_k-\overline{y_k})(\hat{y}_k-\overline{\hat{y}_k})}{\sqrt{\sum (y_k-\overline{y_k})^2 \sum (\hat{y_k}-\overline{\hat{y_k}})^2}}
\end{gather*}
where $\overline{y_k}$ and $\overline{\hat{y_k}}$ are the mean values of the ground truth and the model's predictions, respectively.

\noindent\par\textbf{Calibration of ODE.} 
The target variables for COVID-19 and flu forecasting are different. In COVID-19, we want to calibrate/predict using COVID-associated mortality-- which is more reliable than confirmed cases~\cite{cramer_evaluation_2021}--, while in flu we use influenza-like-illness (ILI) counts, which is collected by the CDC.
ILI measures the percentage of healthcare seekers who exhibit influenza-like-illness symptoms, defined as "fever (temperature of 100°F/37.8°C or greater) and a cough and/or a sore throat without a known cause other than influenza"~\cite{cdc_flu_surveillance}.
Data on COVID-19 mortality can be directly associate to our SEIRM. However, to connect ILI to the SIS model, we need to estimate it based on the available variables.  For this, we use $ \text{ILI \%} = \frac{\beta(t) I S}{N} / (N \cdot \text{OR}) $, where $\text{OR}$ is the outpatients ratio, which is the proportion of the population that are outpatients in a given day. We set $\text{OR}$ based on CDC flu facts~\footnote{\url{cdc.gov/flu/about/keyfacts.htm}}.

\begin{table*}[t!]
\centering
\caption{
\ourmethod vs standard data-driven baselines. Top model per column is in bold.
} 
\resizebox{0.70\textwidth}{!}{
\begin{tabular}{@{\extracolsep{8pt}}lccccccc}
\toprule

{} & \multicolumn{3}{c}{}  & \multicolumn{3}{c}{} & {\textbf{Trend}}\\
{} & \multicolumn{3}{c}{\textbf{Short-term (1-4 wks)}}  & \multicolumn{3}{c}{\textbf{Long-term (5-8 wks)}} & {\textbf{correlation}}\\
 \cmidrule{2-4} 
 \cmidrule{5-7} 
 \cmidrule{8-8} 
 \textbf{Model} & \textbf{NR1} & \textbf{NR2} & \textbf{ND} & \textbf{NR1} & \textbf{NR2} & \textbf{ND} & \textbf{PC} \\ 
 \midrule \midrule
 \multicolumn{8}{c}{\textbf{Task 1: COVID-19 Forecasting (US National + 47 states)}} \\ 
\midrule \midrule
\textbf{\ourmethod (ours)}  & \textbf{0.54}	& \textbf{0.24} &	\textbf{0.38} &	\textbf{0.85} &	\textbf{0.37} &	\textbf{0.66} &	\textbf{0.46} \\
COVID Forecast Hub baseline & 
0.80 & 0.36 & 0.62 & 0.93 & 0.40 & 0.74 & NaN \\
Autoregressive model		
& 0.78 & 0.35 & 0.60 & 1.25 & 0.55 & 0.92 & 0.03 \\
Lasso model w/ features	
& 0.87 & 0.39 & 0.66 & 0.89 & 0.38 & 0.67 & -0.18\\
\midrule \midrule
\multicolumn{8}{c}{\textbf{Task 2: Influenza Forecasting (10 HHS regions)}} \\ \midrule \midrule
\textbf{\ourmethod (ours)}  & \textbf{0.53} & \textbf{0.27} &	\textbf{0.37} &	1.01 &	\textbf{0.42} &	\textbf{0.73} &	\textbf{0.68} \\ 
COVID Forecast Hub baseline	& 0.74 & 0.39 & 0.69 & 1.15 & 0.49 & 1.10 & NaN \\
Autoregressive model & 0.56 & 0.33 & 0.52 & 0.88 & 0.44 & 0.82 & -0.19 \\
Lasso model w/ features	& 0.60 & 0.36 & 0.55 & \textbf{0.85} & 0.43 & 0.80 & -0.82 \\
\bottomrule
\end{tabular}
}
\label{tab:extra_results}
\end{table*}

\begin{table*}[h!]
\centering
\caption{Hyperparameter sensitivity for \ourmethod over states California, Georgia, Illinois, Texas, and US National.} 
\resizebox{0.7\linewidth}{!}{
\begin{tabular}{@{\extracolsep{4pt}}llccccccc}
   \toprule
{} & {} & \multicolumn{3}{c}{\textbf{Short-term (1-4 wks)}}  & \multicolumn{3}{c}{\textbf{Long-term (5-8 wks)}} & {\textbf{Trend correlation}}\\
 \cmidrule{3-5} 
 \cmidrule{6-8} 
 \cmidrule{9-9} 
 \textbf{Hyperparameter} & \textbf{Value} & \textbf{NR1} & \textbf{NR2} & \textbf{ND} & \textbf{NR1} & \textbf{NR2} & \textbf{ND} & \textbf{PC} \\ 
\midrule
$w^{{\text{ODE}}}$ & 1 & 0.41& 	0.50& 	0.35& 	0.52& 	0.73& 	0.48 & 0.68 \\
 & 10 & 0.38 &	0.46 &	0.32 &	0.51 &	0.70 &	0.47 &	0.72 \\ 
  & 100 & 
  0.34 & 	0.39& 	0.29& 	0.57& 	0.73& 	0.50& 0.55 \\ 
\midrule
$w^{{\text{Transfer}}}$ & 1  & 
0.38 &	0.46 &	0.32 &	0.51 &	0.70 &	0.47 &	0.72 \\ 
& 10 &
0.39 &	0.46 &	0.33 &	0.53 &	0.70 &	0.49 & 0.75	 \\	
& 100  & 0.39 &	0.46 &	0.34 &	0.53 &	0.71 &	0.49 & 0.76	\\	
\bottomrule
\end{tabular}
}
\label{tab:params}
\end{table*}

\section{Extra forecasting results \\(vs. standard data-driven methods)}

While the selection of our baselines is correct to address the main focus of the paper, we compare our method against standard data-driven baselines to contextualize our model's performance with the bigger picture of epidemic forecasting. 
Following~\cite{cramer_evaluation_2021}, the COVID Forecast Hub baseline is a persistence baseline which always predicts the past. We also an autoregressive and a LASSO model which also takes the same features as input. 
As we can see in Table~\ref{tab:extra_results}, our method is the only one with consistently low error (NR1, NR2, ND) and high correlation. 
Note that the persistence baseline (COVID Forecast Hub baseline) predicts a constant trend, thus, correlation is not defined (NaN).

\section{Q5: Hyperparameter sensitivity}
Overall, we found that most hyperparameters are not sensitive. 
The most sensitive ones are the weights for ODE loss for time module $\mathcal{L}^{\text{ODE}}$ and feature module $\mathcal{L}^{\text{ODE-F}}$, and embedding $\mathcal{L}^{\text{Emb}}$ and output losses $\mathcal{L}^{\text{Output}}$. To illustrate this, we vary the loss weight $w^{{\text{ODE}}}$ which is applied to both $\mathcal{L}^{\text{ODE}}$ and $\mathcal{L}^{\text{ODE-F}}$
and another loss weight $w^{{\text{Transfer}}}$ which is applied to both $\mathcal{L}^{\text{Emb}}$ and $\mathcal{L}^{\text{Output}}$.

We analyze hyperparameters sensitivity on five geographical regions in the uptrend of the COVID-19 Delta wave (over 2 months, specifically epidemic weeks 202048 to 202101) which is one of the most difficult to predict due to the unprecedented infectiousness of the Delta variant and shift in human behavior.
In Table~\ref{tab:params}, we can see that~\ourmethod performance is stable across different values of hyperparameters
It is worth noting that it tuning is important to have a correct balance on these losses as increasing one weight on may improve error or correlation but degrade the one facet. As noted in the literature~\cite{wang_understanding_2021}, this is common when working with theory-based constrains and how to best proceed remains an open problem.

\end{document}